\documentclass[runningheads]{llncs}

\usepackage{eccv}

\usepackage{eccvabbrv}

\usepackage{graphicx}
\usepackage{booktabs}

\usepackage{multirow}
\usepackage{colortbl}
\newcommand{\dt}[1]{\fontsize{5pt}{0.1em}\selectfont (#1)}
\newlength\savewidth

\usepackage{pifont}
\usepackage{wrapfig}
\usepackage{float}

\usepackage[accsupp]{axessibility}  

\usepackage{hyperref}

\usepackage{orcidlink}

\begin{document}

\title{Exploring Reliable Matching with Phase Enhancement for Night-time Semantic Segmentation} 

\titlerunning{NightFormer for Night-time Semantic Segmentation}

\author{Yuwen Pan$^{1}$\thanks{Equal contribution} \orcidlink{0000-0002-8665-4886} \and
	Rui Sun$^{1\star}$ \orcidlink{0000-0002-8009-4240} \and
 Naisong Luo$^{1\star}$ \orcidlink{0000-0002-1488-3081} \and
Tianzhu Zhang $^{1,2\,\dagger}$ \orcidlink{0000-0003-1856-9564} \and
	Yongdong Zhang$^{1,3}$\orcidlink{0000-0002-1151-1792} }

\renewcommand{\thefootnote}{\fnsymbol{footnote}}
\footnotetext[4]{Corresponding author}
\authorrunning{Y. Pan et al.}
%
\institute{
University of Science and Technology of China \and
Deep Space Exploration Laboratory \and
State Key Laboratory of Communication Content Cognition, People's Daily Online\\
\email{\{panyw,issunrui,lns6\}@mail.ustc.edu.cn, \{tzzhang,zhyd73\}@ustc.edu.cn}}

\maketitle

\begin{abstract}
  Semantic segmentation of night-time images holds significant importance in computer vision, particularly for applications like night environment perception in autonomous driving systems. However, existing methods tend to parse night-time images from a day-time perspective, leaving the inherent challenges in low-light conditions (such as compromised texture and deceiving matching errors) unexplored. To address these issues, we propose a novel end-to-end optimized approach, named NightFormer, tailored for night-time semantic segmentation, avoiding the conventional practice of forcibly fitting night-time images into day-time distributions. Specifically, we design a pixel-level texture enhancement module to acquire texture-aware features hierarchically with phase enhancement and amplified attention, and an object-level reliable matching module to realize accurate association matching via reliable attention in low-light environments. Extensive experimental results on various challenging benchmarks including NightCity, BDD and Cityscapes demonstrate that our proposed method performs favorably against state-of-the-art night-time semantic segmentation methods.
\end{abstract}

\section{Introduction}
\label{sec:introduction}

\begin{figure}[t]
\centerline{\includegraphics[width=\columnwidth]{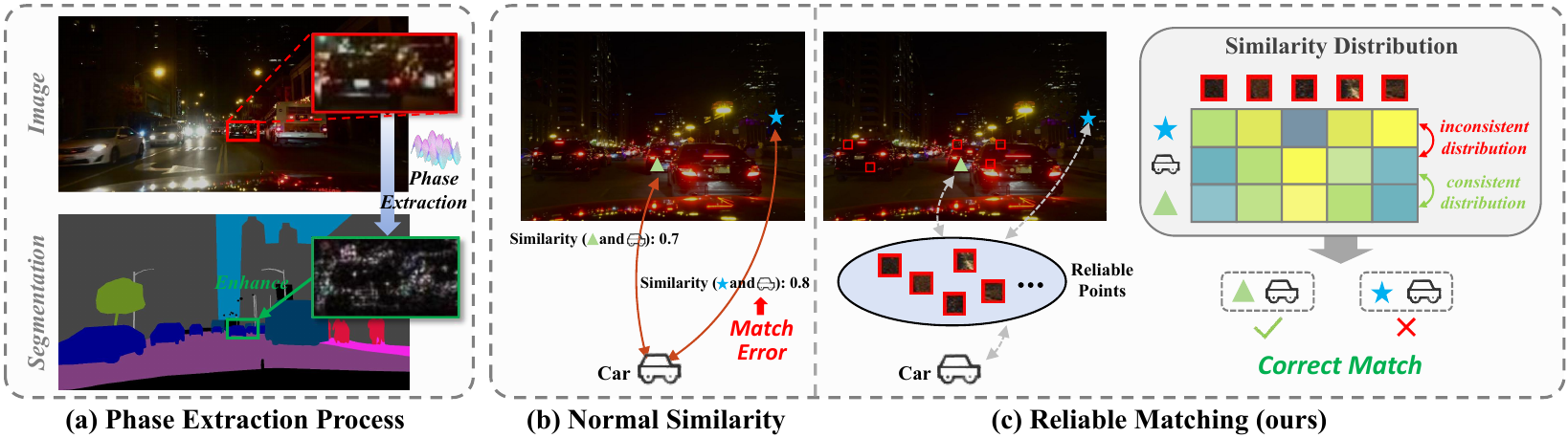}}
\caption{Illustration of our motivation.
(a) 
Due to poor lighting conditions and blurred details at night, 
we utilize Fourier phase decomposition to amplify texture information in night images.
(b)
Normal similarity paradigm tends to directly calculate the similarity between pixels and prototypes,
which may lead to mismatching errors due to deceiving surroundings.
(c)
We propose the reliable attention with adaptively selected reliable points as bridge to calculate similarity rather than direct semantic-pixel matching,
achieving more accurate correlation.
}
\label{fig:fig1}
\end{figure}

Semantic segmentation is a critical computer vision task essential for scene parsing and image processing. 
While existing works predominantly focus on visual perception in daytime scenarios~\cite{yuan2018ocnet, ronneberger2015u, lin2017refinenet}, 
practical applications, 
such as autonomous driving~\cite{li2021deep, schutera2020night}, 
demand robust solutions for challenging night-time environments.
Yet the specific environmental characteristics in night scenes such as low visibility and poor exposure render universally generalized image methods impractical for night-time~\cite{fu2016fusion, tan2021night},
necessitating the development of segmentation methods tailored explicitly for night scenarios.

With significant advances in deep learning (DL), 
DL-based methods have ushered in new research directions for night-time semantic segmentation.
Due to the initial scarcity of night-time image datasets,
early approaches~\cite{wu2021dannet, sakaridis2019guided, song2020nighttime} adopt unsupervised domain adaptation to extend the scene understanding capabilities with the aid of daytime image datasets.
However,
significant challenges arise due to the substantial disparity in illumination and exposure conditions between night-time and day-time images,
leaving a formidable domain gap that poses difficulties in effective bridging.
Recently,
in order to compensate for the lack of night image datasets,
Nightcity~\cite{tan2021night} introduces a comprehensive night-time image dataset,
elevating the research paradigm from the realm of unsupervised methods to a fully-supervised domain.
Several following approaches strategically prioritize the enhancement of night-time images to simulate daytime conditions, 
avoiding a direct confrontation with the inherent challenges posed by night-time scenarios.
Specifically, 
NightLab~\cite{deng2022nightlab} migrates the distribution of night-time images to the daytime domain by collecting nighttime-daytime pairs to supervise the proposed light enhancement module before the segmentation process,
but inevitably requiring additional data.
DTP~\cite{wei2023disentangle} proposes a light-disentangled strategy that decouples the night-time images into light-invariant and light-specific information and then segments disentangled images,
but achieving an optimal disentanglement strategy demands intricate manual tuning.
These methods, 
which rely on supplementary data~\cite{deng2022nightlab} or manually tuned parameters~\cite{wei2023disentangle} during training, 
lack a guarantee of generalizability across diverse scenarios and datasets,
and the two-stage manner falls short of ensuring end-to-end optimization.
Hence,
the exploration of a comprehensive segmentation paradigm tailored for night-time scenarios becomes imperative to address the existing limitations.

To effectively parse the night-time images,
we aim to adopt an end-to-end optimized methodology from a new perspective, 
avoiding the previous practice of forcibly fitting a distribution from night-time to day-time.
However, 
it is nontrivial to achieve this goal without careful consideration of inherent challenges posed by night-time images: 
(1) 
Owing to the diminished ambient lighting in nocturnal settings, 
the discernibility of textures and other intricate details is considerably compromised (the texture of the car is difficult to identify as shown in the left of Fig.~\ref{fig:fig1}~(a)), 
posing a challenge for the network to capture these crucial visual elements. 
Without accurate texture information, 
effective perception of foregrounds with distinct semantics in night scenes becomes unattainable.
\textit{How to capture texture information in low-light environments} is imperative for comprehensive understanding of night scenes.
(2)
Traditional transformer-based methods~\cite{xie2021segformer, cheng2022masked} normally acquire similarities directly between pixel and learnable prototypes/classifiers.
However,
harsh night scenes present challenges with blurred foreground and background contours in dark/underexposed areas.
Discerning subtle differences becomes arduous,
the normal similarity calculation may lead to erroneous matching (the prototype \textit{car} is deceived by the tricky background \textcolor{blue}{$\star$} in Fig.~\ref{fig:fig1}~(b))
resulting in mismatches during association matching. 
\textit{How to resolve association matching errors in night scenes} to achieve object-level perception is highly expected.

In this paper,
to address the inherent problems in night scenes, 
we propose NightFormer customized for night-time semantic segmentation, 
consisting of a pixel-level texture enhancement module and an object-level reliable matching module. 
(1) 
In the pixel-level texture enhancement module, 
to further capture the image details and texture information, 
we use the phase operation of Fourier transform to focus on the details of the texture in the night scene as shown in Fig.~\ref{fig:fig1}~(a).
To efficiently integrate the extracted texture information into the target features, 
we propose an amplified attention mechanism to hierarchically explore inconspicuous objects of all scales in adverse conditions. 
(2) 
In the object-level reliable matching module, 
we first propose a set of prototypes to capture semantic information in the night scenes. 
And in order to realize accurate association matching in low-light environments, 
we further design a reliable attention mechanism to adaptively select the dependable key points as the medium, 
and use them as the bridge to obtain their relationship with the prototypes and the similarity distribution of night-time image features, 
since similar pixels and prototypes exist in a near-consistent distribution as shown in Fig.~\ref{fig:fig1}~(c). 
In this way,
we are able to achieve effective parsing and segmentation of the night scene in both pixel and object-level manners with desirable results.

To sum up, 
our contributions can be summarized as follows:
\begin{itemize}
    \item We propose NightFormer specially designed for semantic segmentation in night-time scenes, offering a novel perspective without forcing the night-time images to fit the distribution of the day-time domain.
    \item We design a pixel-level texture enhancement module to acquire phase-enhanced features via hierarchical amplified attention,
    and an object-level reliable matching module to realize accurate association matching via reliable attention in low-light environments.
    \item Extensive experimental results on various challenging benchmarks including NightCity, BDD and Cityscapes demonstrate that our proposed method performs favorably against state-of-the-art night-time semantic segmentation methods.
\end{itemize}

\section{Related Work}
\label{sec:related work}

\subsection{Semantic Segmentation}
Semantic segmentation is a fundamental task in computer vision aimed at understanding and parsing visual information at the pixel level~\cite{chen2018encoder,xiao2018unified,sun2021lesion,wang2022adaptive,luo2023camouflaged,wang2023rethinking,mai2023dualrel,sun2023appearance,mai2024pay,wang2024image,mai2024rankmatch,xiongaggregation}.
With the development of deep learning,
Fully Convolutional Networks (FCNs~\cite{long2015fully}) emerged as pioneering architectures, 
allowing end-to-end pixel-wise classification. 
Subsequent models, 
including U-Net~\cite{ronneberger2015u} and SegNet~\cite{badrinarayanan2017segnet}, 
introduced innovations like skip connections and encoder-decoder architectures to enhance segmentation accuracy.
In well-illuminated scenarios, 
state-of-the-art methods~\cite{sun2023alignment,wang2023focus,luo2024electron,sun20221st} leverage powerful deep neural networks such as DeepLab~\cite{chen2018encoder} and PSPNet~\cite{zhao2017pyramid}. 
These models benefit from large-scale datasets, 
enabling effective feature learning and context aggregation for precise semantic understanding in daytime environments.
Though achieving promising results, 
these methods cannot generalize well in night scenarios due to challenging illumination environments.
In this paper, 
we propose a customized semantic segmentation network to address inherent issues in night-time images.

\subsection{Night-time Semantic Segmentation}
Night-time semantic segmentation, 
a critical aspect in computer vision for applications such as autonomous driving in low-light conditions, 
has garnered significant attention. 
Early works~\cite{sakaridis2019guided, song2020nighttime, wu2021dannet, gao2022cross, sakaridis2020map} tend to bridge the gap between daytime and night-time conditions,
leveraging unsupervised domain adaptation, with daytime image datasets to enhance the segmentation capabilities in night scenes. 
Recently,
with the introduction of a large night-time dataset proposed by NightCity~\cite{tan2021night},
the task of night-time semantic segmentation has shifted from domain adaptation to a fully-supervised approach.
NightLab~\cite{deng2022nightlab} introduces a dual-level architecture using a light-enhanced module to output augmented images and a hard region detector to optimize difficult part recognition,
prompting the vision system but with redundant modules.
Furthermore, 
DTP~\cite{wei2023disentangle} parses the night scenes by adopting a lighting-disentangled strategy to enhance existing day-time methods for night-time segmentation,
which alleviates the limitations of tough illumination conditions at night time to a certain extent.
Notably,
existing methods tend to map or decouple night-time images into the daytime domain before generating the results with a universal segmentation architecture,
leaving the inherent challenges in night scenes unexplored.
In this paper,
we aim to address the task of night-time semantic segmentation essentially by enhancing texture details and achieving reliable matching in low-light conditions.

\section{Method}
\label{sec:method}
In this section,
we first present the overview of the proposed NightFormer in Sec.~\ref{method:overview}.
Then we describe the details of the pixel-level texture enhancement module in Sec.~\ref{method:enhancement} and the object-level reliable matching module in Sec.~\ref{method:reliable}.
Finally,
in Sec.~\ref{method:training},
the training and inference procedure are discussed.

\begin{figure}[tbp]
\centering
\includegraphics[width=\columnwidth]{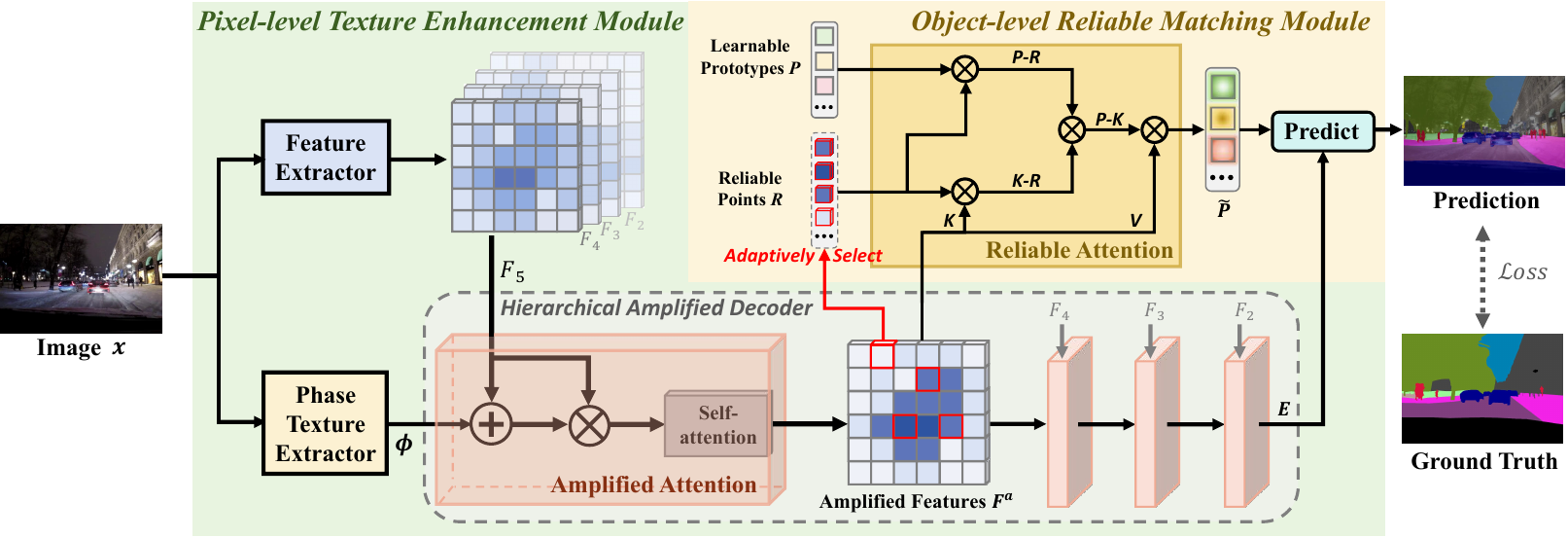}
\caption{
Framework of our proposed NightFormer.
It includes a pixel-level texture enhancement module (Sec.~\ref{method:enhancement}) to hierarchically aggregate phase texture into target information with amplified attention and an object-level reliable matching module (Sec.~\ref{method:reliable}) to realize accurate matching between prototypes and pixels with reliable attention.
}  
\label{fig:method}
\end{figure}

\subsection{Overview}
\label{method:overview}
As shown in Fig.~\ref{fig:method},
given an input image $x \in \mathbb{R}^{H\times{W}\times{3}}$,
where $H$ and $W$ refer to the height and width of the input,
respectively.
The pixel-level texture enhancement module extracts the hierarchical pixel embeddings with an image encoder and the texture-aware features with a Fouier-based encoder,
the following hierarchical amplified decoder combines them later through amplified attention of multiple scales.
The enhanced features $F^a$ are then fed into the object-level reliable matching module to realize accurate perception with reliable points and update the learnable prototypes by empowering them with semantic-aware abilities.
Finally,
the updated prototypes are interacted with restored high-level features to generate the final segmentation result.

\subsection{Pixel-level Texture Enhancement Module}
\label{method:enhancement}
In order to disentangle pixel-level texture and target information fused in the pixel feature,
we first model the representation of texture details
by exploring Fourier frequency domain.
Then we amplify the latent texture information in the acquired pixel-level features with the hierarchical amplified decoder.

\subsubsection{Phase Texture Extraction.}
Existing methods leveraging Fourier Transform typically focus on either optimizing computational costs~\cite{chi2020fast} or transferring styles between different domains~\cite{yang2020fda}. 
In contrast, 
we aim to explore the potential benefits of the Fourier spectrum in segmentation.
In the frequency domain,
it is known that the phase component of Fourier spectrum preserves high-level statistics information,
which contains essential information about the image structure and texture~\cite{yang2020fda}.
Therefore, we can utilize the phase of Fourier spectrum to enhance blurred /compromised details in night-time images.
Specifically, 
in the phase texture extractor,
given a night-time image $x \in \mathbb{R}^{H\times W \times 3}$,
we apply a two-dimensional Fourier transform $\mathcal{F}(x)$ as:
\begin{equation}
\mathcal{F}(x)_{u, v}=\sum_{i=0}^{H-1} \sum_{j=0}^{W-1} x_{i, j} e^{-J 2 \pi\left(\frac{ui}{H} +\frac{vj}{W} \right)},
\end{equation}
where $J$ refers to the imaginary unit.
Then we can acquire the corresponding amplitude $\mathcal{A}$ and phase $\Phi$ as:
\begin{equation}
\mathcal{A}(x)_{u, v}=\left| \mathcal{F}(x)_{u, v} \right|,
\end{equation}
\begin{equation}
\Phi(x)_{u, v}=\arg(\mathcal{F}(x)_{u, v})=\arctan\left[\frac{\text{{Im}}(\mathcal{F}(x)_{u, v})}{\text{{Re}}(\mathcal{F}(x)_{u, v})}\right],
\end{equation}
where $\text{{Im}}$ and $\text{{Re}}$ represent the imaginary and real part of $\mathcal{F}(x)$, respectively.
In this way,
to finally generate our phase texture map,
we fix the amplitude to an average constant $c^a$ and
apply inverse Fourier transform to acquire the phase texture map as
\begin{equation}
    \overline{\Phi}(x) = \mathcal{F}^{-1}[\Phi(x)_{u,v} e^{-Jc^a}],
\end{equation}
where $\mathcal{F}^{-1}$ denotes the inverse Fourier transformation.
Then,
we can obtain the phase characteristic $\phi$ through a light-weight encoder (\textit{e.g.}, ResNet-18~\cite{he2016deep}),
preparing for the following amplification of extracted pixel-level features.

\subsubsection{Hierarchical Amplified Decoder.}

To acquire fine-grained target information with more texture details,
we first obtain the features from different stages of the backbone network $\{F_5,\,F_4,\,F_3,\,F_2\}$,
and then generate the corresponding amplified pixel-level features through the amplified attention mechanism at each stage.

Inspired by the success of Transformer architecture in discovering local regions, 
we further explore the potential of attention mechanism in enhancing phase texture information for night-time segmentation.
Given an image feature $\mathbf{F} \in \mathbb{R}^{h \times w \times c}$ extracted from
the feature extractor,
we amplify the phase characteristics (\textit{i.e.}, texture information) $\phi$
to better restore compromised details in undesirable exposure conditions
through
a novel \textbf{amplified attention} mechanism.
First, 
we employ two convolution layers to map $\mathbf{F}$ and $\phi$
to the same dimension $C$,
then we can obtain $\mathbf{\overline{F}} \in \mathbb{R}^{h\times w \times C}$ and $\overline{\phi} \in \mathbb{R}^{h\times w \times C}$, respectively.
Rather than affinities as vanilla transformer attention mechanism~\cite{vaswani2017attention},
we custom design the amplified attention to integrate phase texture into target features in a finer manner,
prompting the vision system to highlight salient regions in night-time images.
Specifically, 
we acquire the \textit{amplified map} $\mathbf{A}\in \mathbb{R}^{h\times w}$ from pixel-level features and phase characteristic as:
\begin{equation}
    \mathbf{A}_{i,j} = \sum^{C}_{c=1}(\mathbf{\overline{F}}_{i,j,c} + \overline{\phi}_{i,j,c})^2,
\end{equation}
where $i,j$, and $c$ are the index of height, width, and channel, respectively.
Finally,
we can get the amplified pixel features $\mathbf{F}^a$ by weighting the \textit{amplified map} $\mathbf{A}$ to original image features in a pixel-wise manner as:
\begin{equation}
    \mathbf{F}^a = \mathbf{F} \circ \mathbf{A},
\end{equation}
where $\circ$ denotes the element-wise product.
Besides, we use a self-attention layer
for further aggregating target information across different pixels.

The amplification process is repeated with upsampling operations until the final high-resolution feature $\mathbf{E}$ is obtained.
Compared with boundary detection, 
utilizing texture information obtained from phase can more effectively mine the detailed information in night scenes, 
guiding the model to parse these scenes in a unified and flexible manner.
In this way,
the hierarchical amplified decoder is able to capture texture information ranging from coarse to fine details,
beneficial for accurately delineating objects and regions with varying complexities.
Experiments show that this multi-level amplified design contributes considerable performance gain to final semantic segmentation as shown in Tab.~\ref{tab:hierarchical}.

\subsection{Object-level Reliable Matching Module}
\label{method:reliable}

To effectively aggregate target information with different semantics,
we learn a set of prototypes $\mathbf{P} = \{\boldsymbol{p}_n\}^N_{n=1}$,
where $\boldsymbol{p}_n \in \mathbb{R}^{1\times L}$ and N denotes the number of prototypes.
The prototypes,
\textit{i.e.},
learnable query vectors,
can absorb class-wise knowledge via cross-attention in a dynamic manner~\cite{cheng2021per,cheng2022masked,pan2023adaptive}.
They can evolve into a compact and distinct representation for each semantic class
and
serve as anchors around vision features,
facilitating more effective aggregation.
For each layer in the object-level reliable matching module,
these learnable prototypes are first fed into a self-attention layer,
where all keys, queries and values arise from initial prototypes
to incorporate the local context in night-time images.
As the surrounding backgrounds can be deceptive due to the tough illumination environments at night,
the direct similarity calculation is susceptible to background pixel interference,
resulting in erroneous segmentation.
Thus we design a novel reliable attention mechanism to find related reliable points as bridge to acquire more accurate correlations.

\noindent\textbf{Reliable Attention Mechanism.}
Given the amplified feature $\mathbf{F}^a$ derived from the pixel-level texture enhancement module,
the queries arise from the prototypes $\mathbf{P}$, and
keys and values arise from the input features $\mathbf{\tilde{F}}^a = [\boldsymbol{f}_1^a;\boldsymbol{f}_2^a;...;\boldsymbol{f}_{hw}^a]$ (flattened $\mathbf{F}^a$). 
Formally,
\begin{equation}
    \label{eq:weight}
    \mathbf{Q}_{n}=\boldsymbol{p}_n \mathbf{W}^{Q}, \mathbf{K}_{m}=\boldsymbol{f}_m^a \mathbf{W}^{K}, \mathbf{V}_{m}=\boldsymbol{f}_m^a \mathbf{W}^{V},
\end{equation}
where $n \in [1, \ldots, N]$ ,
$m \in 1,2,...,hw$ and $\mathbf{W}^{Q} \in \mathbb{R}^{C \times C_k}$,
$\mathbf{W}^{K} \in \mathbb{R}^{C \times C_k}$,
$\mathbf{W}^{V} \in \mathbb{R}^{C \times C_v}$ are linear projections.
We then can obtain the similarity between $\mathbf{Q}_{n}$ and $\mathbf{K}_{m}$ as:
\begin{equation}
sim_{n, m}=\frac{\exp \left(\beta_{n, m}\right)}{\sum_{m=1}^{h w} \exp \left(\beta_{n, m}\right)}, \beta_{n, m}=\frac{\mathbf{Q}_{n} \mathbf{K}_{m}^\top}{\sqrt{C_{k}}}.
\label{eq:cross}
\end{equation}

Direct similarity calculation is unreliable due to the high similarity between foreground pixels with different semantics in night scenes, 
especially in underexposed regions.
We aim to find a reliable medium that can be used as a basis to build a matching bridge between prototypes and pixels. 
The reliable score for each pixel can be obtained via the weighted sum over all similarities as:
\begin{equation}
    score_m = \sum^{N}_{n=1}sim_{n, m}, m\in 1,2,...,hw,
\end{equation}
where the top-$K$ pixels are selected with the largest correlations of semantics
to be reliable points $\mathbf{R}$.
The sum of similarities between a pixel and all prototypes $score_m$ can be regarded as an aggregated measure of confidence/reliability. 
Each similarity score indicates the degree to which a pixel's feature vector aligns with a prototype. 
By summing these similarities, 
we capture the overall confidence/reliable score (how well a pixel's feature vector is semantically consistent with multiple prototypes),
which implies that the pixel tends to be well-represented within the semantic space spanned by the prototypes, 
making it a reliable candidate for accurate discrimination.
The corresponding features of reliable points $\mathbf{R}$ can be denoted as $\textbf{F}^R = \{\boldsymbol{f}_n^R\}^K_{k=1}$.
Then we calculate the prototype-reliable similarity and the pixel-reliable similarity respectively as same as Eq.(\ref{eq:cross}):
\begin{equation}
\begin{aligned}
    & Sim^{q}_n = \operatorname{softmax}(\frac{(\boldsymbol{p}_n \mathbf{W}^{Q})(\textbf{F}^R \mathbf{W}^{K})^\top}{\sqrt{C_{k}}}),  \\
    & Sim^{k}_m = \operatorname{softmax}(\frac{(\boldsymbol{f}_m^a \mathbf{W}^{Q})(\textbf{F}^R \mathbf{W}^{K})^\top}{\sqrt{C_{k}}}).
\end{aligned}
\end{equation}

And then,
we can obtain the similarity between prototypes and pixels with reliable points as:
%
\begin{equation}
    Sim^{qk}_{n,m} = Sim^{q}_n(Sim^{k}_m)^\top,
\end{equation}
which is used to acquire more accurate correlations.
Finally,
the updated prototypes can be acquired by blending values with the reliable similarity $Sim^{qk}_{n,m}$ as:
\begin{equation}
    \boldsymbol{\tilde{p}}_n = \sum^{hw}_{m=1}Sim^{qk}_{n,m}\mathbf{V}_{m},
\end{equation}
and following general transformer pipeline~\cite{vaswani2017attention}, we equip updated prototypes with self-attention and FFN at the output of the reliable attention.
In this way,
the learnable prototypes $\mathbf{P}$ are modified in the object-level reliable matching module via reliable attention and finally evolve into reliable classifiers $\mathbf{\tilde{P}}$.

\subsection{Training and Inference}
\label{method:training}
With the final high-resolution features $\mathbf{E} \in \mathbb{R}^{\frac{H}{4}\times 
\frac{W}{4} \times{C}}$ and the learned prototypes  $\mathbf{\tilde{P}} \in \mathbb{R}^{N \times {C}}$ as classifiers,
we can finally obtain the segmentation map as:
\begin{equation}
    \label{eq:seg}
    \mathbf{M} = \mathbf{E} \times \mathbf{\tilde{P}}^\top.
\end{equation}

For better training our network,
we use the conventionally used loss paradigm~\cite{deng2022nightlab, tan2021night},
including the dice loss and binary cross-entropy loss to supervise mask prediction and the cross-entropy loss for mask recognition.

\section{Experiments}
\label{sec:experiments}
In this section,
we will first introduce the datasets used in our work in Sec.~\ref{exp:dataset}.
And the implementation details are shown in Sec.~\ref{exp:implementation}.
In Sec.~\ref{exp:metric},
we illustrate the specific metric for better evaluation of our method.
Then we further analyze the main results including quantitative evaluations and qualitative results in Sec.~\ref{exp:main}.
Finally,
we ablate the effectiveness of our method in Sec.~\ref{exp:ablation}.

\begin{table}[t!]
\centering
\caption{Comparisons of existing night-time semantic segmentation methods and general segmentation approaches on the NightCity~\cite{tan2021night} and NightCity-fine~\cite{wei2023disentangle} datasets.
The best results are shown in \textbf{bold}.
}
\resizebox{1\columnwidth}{!}{
\centering
\renewcommand{\arraystretch}{1} 
\setlength\tabcolsep{8pt}
\begin{tabular}{l|c|c|cc}
\toprule[1pt]
Method             & \multicolumn{1}{l|}{Backbone} & Parameters                    & NightCity & NightCity-fine \\ \hline
NightCity~\cite{tan2021night}          &   ResNet101     & 84.6M                         & 51.8      & 55.9           \\
PSPNet~\cite{zhao2017pyramid}             &   ResNet101   & 88.3M  & 46.3      & 49.5           \\
DeeplabV3+~\cite{chen2018encoder}         &    ResNet101   & 60.1M                         & 54.7      & 58.8           \\
DANet~\cite{fu2019dual}              &    ResNet101    & 76.3M  & 56.0      & 59.3           \\
NightLab~\cite{deng2022nightlab}           &    ResNet101    & 98.5M  & 55.9      & 62.3           \\
DTP~\cite{wei2023disentangle}                &    ResNet101  & 63.9M                         & 57.6      & 60.4           \\ \rowcolor{gray!15}
\textbf{NightFormer (ours)} & ResNet101    &     58.4M                          &  \textbf{59.8}\textcolor{blue}{\dt{$\uparrow$2.2}}         &  \textbf{62.8}\textcolor{blue}{\dt{$\uparrow$2.4}}              \\ \hline
UPerNet~\cite{xiao2018unified}            &    Swin-Base             & 102.5M & 57.7      & 60.5           \\
UPer-Swin~\cite{liu2021swin}          &    Swin-Base & 119.9M                        & 58.4      & 61.1           \\
NightLab~\cite{deng2022nightlab}           &    Swin-Base        & 242.4M                        & 59.8      & 62.3           \\
DTP~\cite{wei2023disentangle}                &    Swin-Base    & 122.5M                        & 61.2      & 64.2           \\ \rowcolor{gray!15}
\textbf{NightFormer (ours)} & Swin-Base  &    111.9M                           &  \textbf{63.5}\textcolor{blue}{\dt{$\uparrow$2.3}}         &  \textbf{65.9}\textcolor{blue}{\dt{$\uparrow$1.7}}              \\ 
\bottomrule[1pt]
\end{tabular}
}
\label{tab:nightcity}
\end{table}

\subsection{Dataset}
\label{exp:dataset}
Following the conventional practice~\cite{tan2021night, wei2023disentangle},
there are 4 datasets included to evaluate the night-time semantic segmentation performance of our method:
NightCity~\cite{tan2021night},
NightCity-fine~\cite{wei2023disentangle},
CityScapes~\cite{cordts2016cityscapes},
and BDD100K~\cite{yu2020bdd100k}.

\noindent\textbf{NightCity}~\cite{tan2021night}
is a dataset containing 4,297 real night-time images,
divided into 2,998 train images and 1,299 val images with pixel-level semantic annotations.
The labels align with Cityscapes~\cite{cordts2016cityscapes} including 19 classes.

\noindent\textbf{NightCity-fine}~\cite{wei2023disentangle}
is a refined version of NightCity with the same images.
It identifies mislabeled validation images and re-annotates more accurate labels with the assistance of human labelers for better evaluation.

\noindent\textbf{CityScapes}~\cite{cordts2016cityscapes}
is a commonly used benchmark dataset for semantic segmentation tasks.
Please note that urban images in this dataset are mostly in daytime scenes.
Following the previous setting~\cite{tan2021night,wei2023disentangle},
we only use the training set of CityScapes to aid the training procedure as a reference to the validity of our method as shown in Tab.~\ref{tab:n+c}. 

\noindent\textbf{BDD100K}~\cite{yu2020bdd100k}
is a large-scale, diverse driving dataset with different weather conditions including night-time (B-N) and day-time (B-D).
For our supplementary experiments,
we only utilize the night images in BDD100K as BDD100K-night following the setting in~\cite{deng2022nightlab} with a validation set of 58 images,
while the other 7000 images are used for training due to the scarcity of night-time images.

\subsection{Implementation Details}
\label{exp:implementation}
We adopt Pytorch~\cite{paszke2019pytorch} and Detectron2~\cite{wu2019detectron2} to implement the proposed method. 
4 NVIDIA GeForce RTX 3090 GPUs are used for training. 
We consider both ResNet101 and Swin-Base image backbones following~\cite{deng2022nightlab,wei2023disentangle} for better evaluation.
The extractor of phase texture is ResNet-18~\cite{he2016deep},
which is a light-weight CNN encoder.
During the training stage, 
our model is trained with a batch size of 16, 
using the Adam optimizer~\cite{loshchilov2017decoupled} with an initial learning rate of 0.0001 for the first 60,000 iterations
and 0.00001 for the last 20,000 iterations.
The input image is resized to the resolution of $512 * 1024$.
We set the number of learnable prototypes as $N = 32$,
and the number of reliable points as $K = 40$.
We ablate the effects of these super-parameters in detail in our ablation studies as shown in Fig.~\ref{fig:ablation}.

 \begin{table}[!t]
    \centering
	\begin{minipage}[t]{0.47\hsize} 
   \renewcommand\arraystretch{1.065} 
   \setlength\tabcolsep{3pt}
   \centering
   \caption{Comparisons of results on the NightCity~\cite{tan2021night} and CityScapes~\cite{cordts2016cityscapes} datasets.
Please note that the training procedure (NightCity\&CityScapes) includes both training sets.
}
   \resizebox{1.0\hsize}{!}{
\begin{tabular}{l|c|cc}
\toprule[1.5pt]
\multicolumn{1}{c|}{\multirow{2}{*}{Method}} & \multirow{2}{*}{Backbone}  & \multicolumn{2}{c}{Trained on NightCity\&CityScapes} \\ \cline{3-4} 
\multicolumn{1}{c|}{}                        &                            & NightCity                & CityScapes               \\ \hline
NightCity~\cite{tan2021night}                                    & ResNet101 & 53.9                     & 76.9                     \\
DeeplabV3+~\cite{chen2018encoder}                                   & ResNet101                           & 59.0                     & 73.6                     \\
DTP~\cite{wei2023disentangle}                                          & ResNet101                           & 59.9                     & 75.2                     \\ \rowcolor{gray!15}
\textbf{NightFormer (ours)}                  & ResNet101                           & \textbf{62.2}\textcolor{blue}{\dt{$\uparrow$2.3}}            & \textbf{80.3}\textcolor{blue}{\dt{$\uparrow$5.1}}            \\ \hline
UPer-Swin~\cite{liu2021swin}                                    & Swin-Base & 59.7                     & 76.0                       \\
NightLab~\cite{deng2022nightlab}                                     & Swin-Base                           & 60.2                     & 77.1                     \\
DTP~\cite{wei2023disentangle}                                          & Swin-Base                           & 63.3                     & 78.3                     \\ \rowcolor{gray!15}
\textbf{NightFormer (ours)}                  & Swin-Base                           & \textbf{65.4}\textcolor{blue}{\dt{$\uparrow$2.1}}            & \textbf{82.1}\textcolor{blue}{\dt{$\uparrow$3.8}}            \\ 
\bottomrule[1.5pt]
\end{tabular}
}
  \label{tab:n+c}
  \end{minipage} 
  \begin{minipage}[t]{0.49\hsize} 
   \renewcommand\arraystretch{1.1} 
   \setlength\tabcolsep{3.5pt}
   \centering
   \caption{Comparisons of results on the BDD100K dataset of different training procedures.
B-N denotes the BDD100K-night~\cite{deng2022nightlab} training set,
B-N$\&$B-D denotes the whole BDD100K~\cite{yu2020bdd100k} training set.
}
   \resizebox{1.0\hsize}{!}{
\begin{tabular}{l|l|cc}
\toprule[1.5pt]
\multicolumn{1}{c|}{\multirow{2}{*}{Method}} & \multicolumn{1}{c|}{\multirow{2}{*}{Backbone}} &  Trained on B-N & Trained on B-N\&B-D   \\ \cline{3-4} 
\multicolumn{1}{c|}{}                        & \multicolumn{1}{c|}{}                          & \multicolumn{2}{c}{BDD100K-night} \\ \hline
NightCity~\cite{tan2021night}                                    & ResNet101                    & 28.4           & 39.7                \\
DeeplabV3+~\cite{chen2018encoder}                                   & ResNet101                                               & 30.1           & 43.4                \\
NightLab~\cite{deng2022nightlab}                                     & ResNet101                                               & 31.3           & 45.1                \\ \rowcolor{gray!15}
\textbf{NightFormer (ours)}                  & ResNet101                                               & \textbf{32.8}\textcolor{blue}{\dt{$\uparrow$1.5}}  & \textbf{47.5}\textcolor{blue}{\dt{$\uparrow$2.4}}       \\ \hline
UPer-Swin~\cite{liu2021swin}                                    & Swin-Base                     & 31.7           & 48.0                \\
NightLab~\cite{deng2022nightlab}                                     & Swin-Base                                               & 35.4           & 50.4                \\
DTP~\cite{wei2023disentangle}                                          & Swin-Base                                               & 36.9           & 53.1                \\ \rowcolor{gray!15}
\textbf{NightFormer (ours)}                  & Swin-Base                                               & \textbf{38.2}\textcolor{blue}{\dt{$\uparrow$1.3}}  & \textbf{55.4}\textcolor{blue}{\dt{$\uparrow$2.3}}       \\ 
\bottomrule[1.5pt]
\end{tabular}
}
\label{tab:bdd}
  \end{minipage} 
\end{table}

\subsection{Metric}
\label{exp:metric}

For a fair comparison,
we adopt the metric of mean intersection over union (mIOU),
which is a commonly used evaluation metric for image segmentation tasks.
It measures the similarity between the predicted segmentation mask and the ground truth mask
by computing the ratio of the intersection area to the union area of the two masks for each object class.

\subsection{Main Results}
\label{exp:main}

\subsubsection{Quantitative Evaluations.}

Our method demonstrates superior performance in night-time semantic segmentation,
outperforming state-of-the-art methods, 
as illustrated in Tab.~\ref{tab:nightcity}. 
Evaluation on both the NightCity and NightCity-fine datasets reveals compelling results:
a mIOU of $59.8$ and $62.8$ based on ResNet101, 
and $63.5$ and $65.9$ based on Swin-Base. 
These achievements surpass all existing methods with the same backbone. 
In supplementary experiments, 
as shown in Tab.~\ref{tab:n+c}, 
our NightFormer further enhances scene comprehension when applied to both NightCity and CityScapes datasets. 
It is significant that our method exhibits superior performance not only in night images (NightCity~\cite{tan2021night} dataset) but also demonstrates substantial improvements on the CityScapes dataset across various illumination conditions. 
This underscores our network's capability to assimilate knowledge from both night and day images jointly, 
achieving better information capture from both domains. 
Besides,
as shown in Tab.~\ref{tab:bdd},
whether training on night-only images (B-N) or on a mixed night-day image set (B-N$\&$B-D),
the results on BDD100k-night~\cite{deng2022nightlab} 
further validate our effectiveness.

\begin{figure*}[!t]
    \centering
    \includegraphics[width=\linewidth]{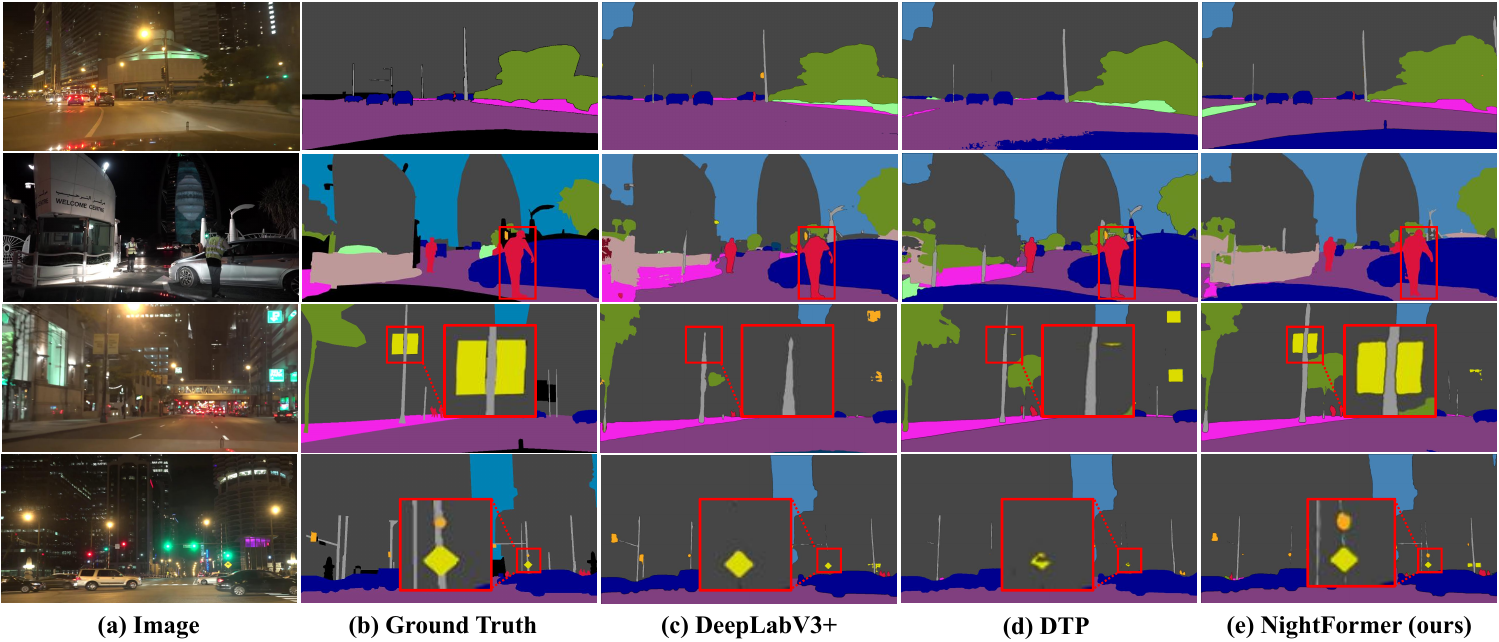}
    \caption{Comparison of qualitative results of our NightFormer and other methods on the NightCity~\cite{tan2021night} dataset.}
    \label{fig:exp-main}
\end{figure*}

\subsubsection{Qualitative Results.}
As shown in Fig.~\ref{fig:exp-main},
NightFormer shows promising segmentation performance in diverse night-time scenes. 
In specific,
our method performs great in most scenarios,
especially in the low-light regions.
Even with some annotation errors in the ground truth set of NightCity~\cite{tan2021night},
our NightFormer still generates promising results for inconspicuous objects such as “traffic sign” and “traffic light” as shown in the last row in Fig.~\ref{fig:exp-main}.
It can also be observed that in the per-class IOU demonstration in Fig.~\ref{fig:perclass}, 
applying our module designs bring significant performance improvements for the categories “traffic sign”, “person” and “truck”.

\subsection{Ablation Study}
\label{exp:ablation}

\noindent{\textbf{Does the Fourier phase extraction contribute to parsing the night-time images?}}
Yes.
As shown in the ablation experiments on main components of our NightFormer in Tab.~\ref{tab:ablation-main},
the incorporation of Fourier phase extraction yields a discernible performance improvement, 
elevating the mIOU from 61.1 to 63.5 on the NightCity dataset. 
The inherent low visibility in night-time images can lead to the loss of texture details in certain targets. 
And the application of Fourier phase extraction proves instrumental in effectively capturing intricate textures. 
\begin{wraptable}{r}{0.55\textwidth} 
\centering
\caption{Ablation of main components on NightCity~\cite{tan2021night} and BDD100k-night~\cite{deng2022nightlab}.}
\resizebox{!}{9mm}{
\begin{tabular}{cc|cc}
\toprule[1.5pt]
\multicolumn{2}{c|}{Main Components} &\multirow{2}{*}{NightCity} &\multirow{2}{*}{BDD100k-night} \\ \cline{1-2}
Phase enhancement & Reliable matching   &  \\ 
\hline
$\times$ & $\times$  & 57.8 & 49.6  \\
\checkmark & $\times$ & 60.4 & 52.1  \\
$\times$ & \checkmark & 61.1 & 53.4  \\ \rowcolor{gray!15}
\checkmark & \checkmark & \textbf{63.5} & \textbf{55.4} \\
\bottomrule[1.5pt]
\end{tabular}
}
\label{tab:ablation-main}
\end{wraptable}
In Fig.~\ref{fig:perclass}, 
we present the detailed per-class IOU comparisons corresponding to Tab.~\ref{tab:ablation-main}. 
Notably, 
our approach exhibits the most significant improvement in recognizing small targets (\textit{e.g.}, \textit{traffic sign} and \textit{traffic light}). 
The phase enhancement promotes the understanding of visual cues, 
leveraging amplified attention to enrich the model's perceptual capabilities.

\noindent{\textbf{Is the hierarchical amplified decoder effective for night-time segmentation?}}
Yes.
As shown in Tab.~\ref{tab:hierarchical},
substantial performance improvements are evident on both datasets when employing multi-level features compared to 
utilizing only the bottom embedding. 
This underscores the effectiveness of our hierarchical fusion design. 
The rationale behind this success lies in the fact that amplified features at different levels are well-suited for capturing textures at diverse scales. 
The incorporation of fused hierarchical features proves instrumental in effectively parsing various types of targets in various night-time scenes.

\noindent{\textbf{Can the reliable attention improve model performance?}}
Yes.
The design of our reliable attention mechanism proves immensely beneficial to the model,
particularly when complemented by phase extraction, 
as shown in Tab.~\ref{tab:ablation-main}. 
Please note that the absence of reliable matching denotes the utilization of a commonly employed cross-attention mechanism.
The adaptive selection of reliable points within this design holds pivotal importance for pixel-level precision in low-visibility or under-exposed regions,
which effectively addresses mismatching issues arising from the conventional direct similarity calculation. 
As shown in Fig.~\ref{fig:attention},
by leveraging the reliable attention mechanism, 
our learnable prototypes effectively cluster high responses in the foreground area while minimizing attention to irrelevant regions. 
In contrast, 
without reliable matching,
prototypes tend to adsorb information from irrelevant areas and activate deceiving regions,
leading to sub-optimal results.
The chosen reliable points also influence the comparison of similarity distributions, 
as illustrated in Fig.~\ref{fig:ablation}~(b).

\begin{center}
\begin{minipage}[!h]{0.41\textwidth}
    \begin{figure}[H] 
        \includegraphics[width=\linewidth]{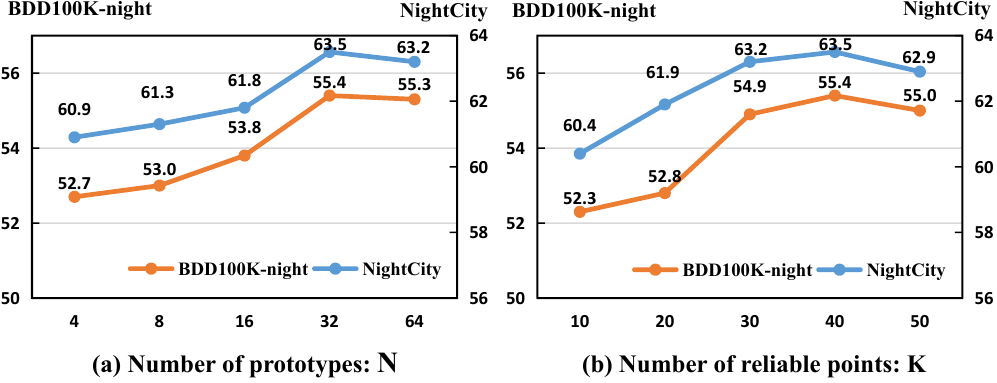}
        \caption{Ablation of $N$ and $K$ on NightCity~\cite{tan2021night} and B-N~\cite{deng2022nightlab}.}
    \label{fig:ablation}
    \end{figure}
\end{minipage}
\begin{minipage}[!t]{0.56\textwidth}
    \begin{figure}[H] 
        \includegraphics[width=\linewidth]{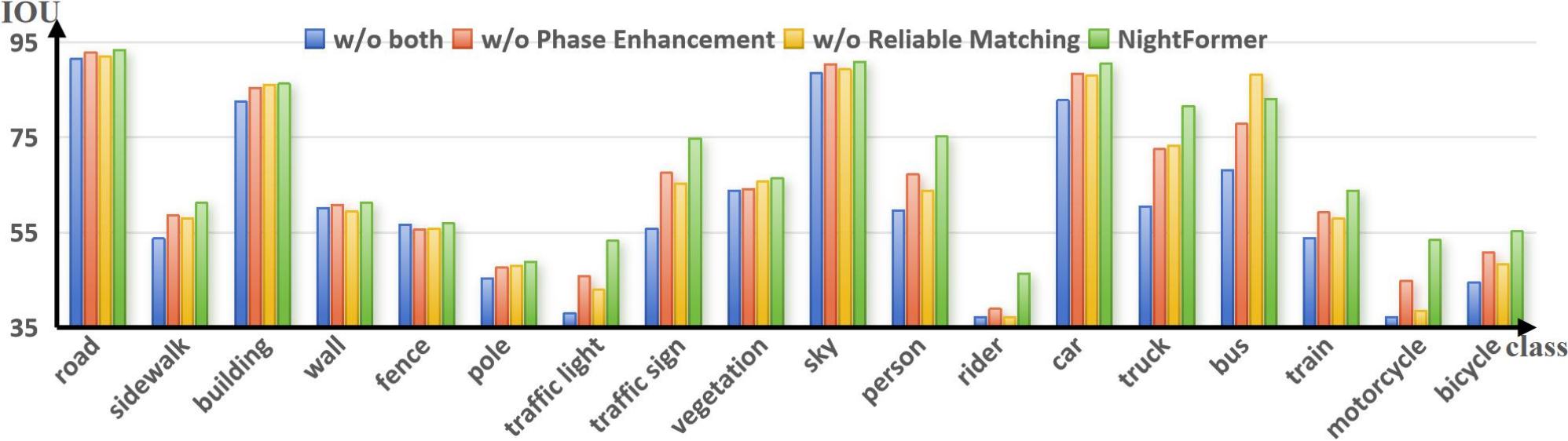}
        \caption{Demonstration of per-class IOU in Tab.~\ref{tab:ablation-main} on the NightCity~\cite{tan2021night} dataset.}
    \label{fig:perclass}
    \end{figure}
\end{minipage}
\end{center}
\begin{table}[!h]
    \centering
	\begin{minipage}[t]{0.47\hsize} 
   \renewcommand\arraystretch{1} 
   \setlength\tabcolsep{3pt}
   \centering
   \caption{Ablation of different hierarchical fusion designs in the fusion layer on both NightCity~\cite{tan2021night} and BDD100K-night~\cite{deng2022nightlab} datasets.
}
   \resizebox{0.9\hsize}{!}{
\begin{tabular}{l|cc}
\toprule[1pt]
Fusion Strategy & NightCity   & BDD100k-night \\ \hline
$\{F_5\}$                        & 61.5 & 53.7 \\
$\{F_5,F_4\}$                        & 61.8 & 53.8 \\
$\{F_5,F_4,F_3\}$                        & 62.9 & 54.6 \\ \rowcolor{gray!10}
$\{F_5,F_4,F_3,F_2\}$                        & \textbf{63.5} & \textbf{55.4} \\ 
\bottomrule[1pt]
\end{tabular}
}
  \label{tab:hierarchical}
  \end{minipage} 
  \hspace{0.005\textwidth}
  \begin{minipage}[t]{0.49\hsize} 
   \renewcommand\arraystretch{1.08} 
   \setlength\tabcolsep{3pt}
   \centering
   \caption{Ablation of different enhancing operations in the pixel-level enhancement module on both NightCity~\cite{tan2021night} and BDD100K-night~\cite{deng2022nightlab} datasets.
}
   \resizebox{0.9\hsize}{!}{
\begin{tabular}{l|cc}
\toprule[1pt]
Enhancing Operation & NightCity   & BDD100k-night \\ \hline
\ding{56}                      & 61.1 & 53.4 \\
Canny Operator                       & 61.9 & 54.3 \\
Sobel Operator                        & 62.1 & 54.4 \\ \rowcolor{gray!10}
Fourier Phase                        & \textbf{63.5} & \textbf{55.4} \\ 
\bottomrule[1pt]
\end{tabular}
}
\label{tab:algorithm}
  \end{minipage} 
\end{table}

\noindent{\textbf{How much does the hyper-parameters affect the model performance?}}
As illustrated in Fig.~\ref{fig:ablation}, 
we conducted ablation experiments focusing on the number of prototypes $N$ and reliable points $K$. 
In the object-level reliable matching module, 
the prototype plays a pivotal role in clustering similar semantics. 
The selection of a specific number of prototypes is crucial for an effective semantic matching process. 
The performance peaks when the number of prototypes is set to 32. 
Deviating towards too many or too few prototypes results in a drop in mIOU performance.
In the reliable attention mechanism, 
the number of reference points, 
denoted as $K$, 
determines how many reliable foreground pixels are selected to establish the correlation between semantics and pixels.
Ablating the number of reference points,
as depicted in Fig.~\ref{fig:ablation}~(b), 
reveals that performance reaches its top when $K = 40$, 
signifying that this number is sufficient for achieving the necessary correction. 
Too many reliable points may overly emphasize model discrimination and include irrelevant background information, ultimately disrupting the similarity distribution.

\begin{figure}[!t]
    \centering
    \includegraphics[width=1\linewidth]{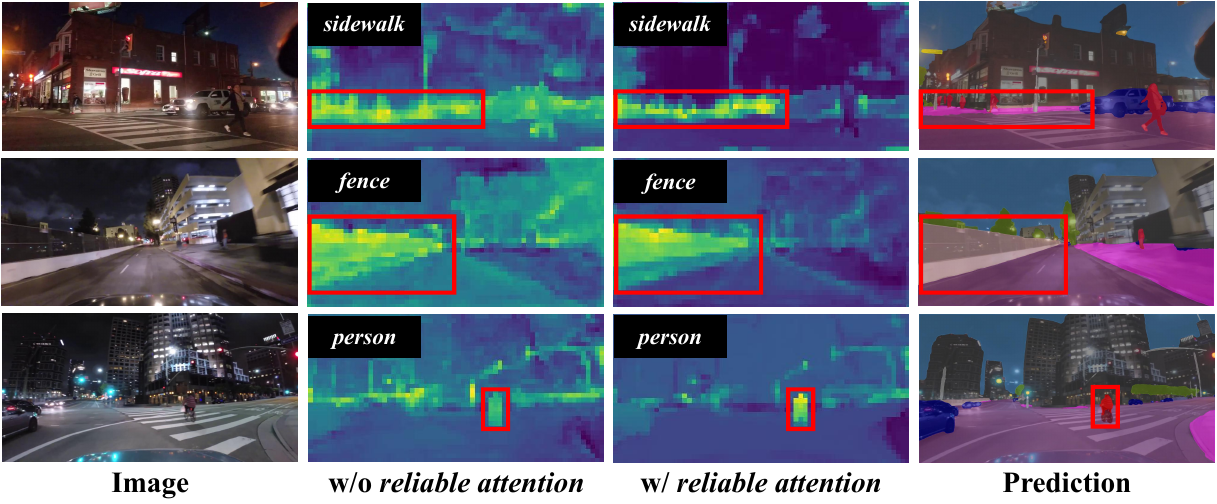}
    \caption{Activation maps of different semantics with and without reliable attention.}
    \label{fig:attention}
\end{figure}

\noindent{\textbf{Is Fourier phase extraction the only way to amplify texture information?}}
In our supplementary experiments, 
we extended our texture enhancement approaches to include other operations, 
such as the Canny operator and Sobel operator. 
As shown in Tab.~\ref{tab:algorithm}, 
when applying traditional algorithms, 
we observed tiny improvement in performance, 
but not as promising as the enhancement achieved through phase extraction. 
We attribute this phenomenon to the fact that certain operations may extract evident contours or textures in a conventional manner. 
However, 
this type of information has already been effectively parsed during the encoding of the original image. 
The repetitive inclusion of contour information does not contribute significantly. 
Besides, 
inherent challenges in night scenes, 
\textit{e.g.}, 
abundant noise and poor lighting, 
result in ambiguous boundary predictions due to deceptive regions. 
In contrast, 
Fourier frequency domain decomposition has the capability to disentangle essential information within different domains for nocturnal images and implicitly enhances visual perception,
making more substantial contributions to amplified pixel-level features.

\section{Conclusion}
\label{sec:conclusion}
In this paper, 
we propose NightFormer for night-time semantic segmentation.
Rather than forcing night-time images to conform to the day-time distribution, 
we aim to efficiently parse night-time scenes by addressing intrinsic challenges such as compromised texture details and mismatching errors.
Specifically, 
we design a pixel-level texture enhancement module that hierarchically aggregates phase texture with amplified attention
and an object-level reliable matching module that accurately matches semantics to pixels using reliable attention. 
Extensive experimental results demonstrate the effectiveness of our proposed NightFormer in night-time semantic segmentation.
Exploring the applicability of our method
in more challenging scenarios
is a promising direction for the future work.

\section*{Acknowledgements}
This work was partially supported by the National Nature Science Foundation of China (Grant 12150007, 62121002, 62071122), and Youth Innovation Promotion Association CAS.

%
%
\bibliographystyle{splncs04}
\bibliography{main}
\end{document}